\documentclass[manuscript, nonacm]{acmart}

\usepackage{amssymb}   
\usepackage{pifont}    
\usepackage{listings}
\usepackage{xcolor}

\lstdefinelanguage{json}{
  basicstyle=\ttfamily\footnotesize,
  numbers=left,
  numberstyle=\tiny,
  stepnumber=1,
  numbersep=8pt,
  showstringspaces=false,
  breaklines=true,
  literate=
   *{0}{{{\color{blue}0}}}{1}
    {1}{{{\color{blue}1}}}{1}
    {2}{{{\color{blue}2}}}{1}
    {3}{{{\color{blue}3}}}{1}
    {4}{{{\color{blue}4}}}{1}
    {5}{{{\color{blue}5}}}{1}
    {6}{{{\color{blue}6}}}{1}
    {7}{{{\color{blue}7}}}{1}
    {8}{{{\color{blue}8}}}{1}
    {9}{{{\color{blue}9}}}{1}
}
\newcommand{\xmark}{\ding{55}} 
\AtBeginDocument{%
  }

\setcopyright{acmlicensed}
\copyrightyear{2018}
\acmYear{2018}
\acmDOI{XXXXXXX.XXXXXXX}
\acmConference[Conference acronym 'XX]{Make sure to enter the correct
  conference title from your rights confirmation email}{June 03--05,
  2018}{Woodstock, NY}
\acmISBN{978-1-4503-XXXX-X/2018/06}

\begin{document}

\title{Symbolically Scaffolded Play: Designing Role-Sensitive Prompts for Generative NPC Dialogue}

\author{Vanessa Figueiredo}
\email{vanessa.figueiredo@uregina.ca}
\orcid{0000-0001-9190-650X}
\author{David Elumeze}
\email{elumezedavid@gmail.com}
\affiliation{%
  \institution{ExplorAI, Department of Computer Science, University of Regina}
  \city{Regina}
  \state{Saskatchewan}
  \country{Canada}
}








\renewcommand{\shortauthors}{Figueiredo \& Elumeze}

\begin{abstract}
Large Language Models (LLMs) promise to transform interactive games by enabling non-player characters (NPCs) to sustain unscripted dialogue. Yet it remains unclear whether constrained prompts actually improve player experience. We investigate this question through \textit{The Interview}, a voice-based detective game powered by GPT-4o. A within-subjects usability study ($N=10$) compared high-constraint (HCP) and low-constraint (LCP) prompts, revealing no reliable experiential differences beyond sensitivity to technical breakdowns. Guided by these findings, we redesigned the HCP into a hybrid \texttt{JSON+RAG} scaffold and conducted a synthetic evaluation with an LLM judge, positioned as an early-stage complement to usability testing. Results uncovered a novel pattern: scaffolding effects were role-dependent: the Interviewer (quest-giver NPC) gained stability, while suspect NPCs lost improvisational believability. These findings overturn the assumption that tighter constraints inherently enhance play. Extending fuzzy–symbolic scaffolding, we introduce \textit{Symbolically Scaffolded Play}, a framework in which symbolic structures are expressed as fuzzy, numerical boundaries that stabilize coherence where needed while preserving improvisation where surprise sustains engagement.
\end{abstract}



\maketitle

\section{Introduction}
Games have always balanced freedom and constraint: from branching dialogue trees to sandbox worlds, designers orchestrate structure while leaving space for improvisation \cite{aylett1999narrative}. Large Language Models (LLMs) intensify this tension by powering non-player characters (NPCs) who can converse in ways that feel unscripted, improvisational, and alive. This shift is often heralded as a revolution, yet it raises a central design question: when do the structural supports we embed in prompts actually matter to players’ experience?

In prior work, we have referred to such structural supports as \emph{scaffolds}: symbolic fuzzy logic schemas combined with a natural language boundary prompt designed to stabilize coherence while still leaving space for variation \cite{figueiredo2025designing, figueiredo2025fuzzy}. We use \emph{scaffolds} to mean structured supports that shape the \emph{conditions of improvisation}. For instance, \texttt{JSON} schemas for turn-taking, role constraints, or retrieval-grounded memory, rather than hard-coded dialogue. In our approach, fuzzy parameters are continuous values in $[0,1]$ that modulate how strongly a scaffold applies at a given moment.

Prior HCI and AI research shows that backend refinements are often invisible unless failures are salient \cite{amershi2019guidelines}. Procedural content generation (PCG) studies likewise grapple with the tension between coherence and variation in emergent narratives, with several studies highlighting trade-offs between structure and freedom \cite{maleki2024pcg, xu2025constraint, whitehead2025conversational}. Recent LLM–game research echoes this pattern: while rigid and highly descriptive prompts can improve technical validity, they rarely guarantee more engaging play \cite{hu2024game, marincioni2024effect, yang2024gpt, gallotta2024consistent}. More rigid and descriptive prompt designs ensure logical flow but suppress spontaneity \cite{lee2023chatgpt, van2021fine}. Yet the gap remains: most evaluations stop at analyzing \emph{logs}—model outputs coded for coherence, factuality, or diversity, without examining how different scaffolds actually shape \emph{lived play} as experienced by players.

We address this gap by introducing \textit{Symbolically Scaffolded Play}, extending fuzzy–symbolic scaffolding \cite{figueiredo2025designing, figueiredo2025fuzzy} into games. We argue that players demand both believability and surprise: rigid scaffolds promise stability but risk dampening improvisation, while looser scaffolds invite variation but risk incoherence. Our studies reveal a paradox: hidden refinements often yield diminishing experiential returns, yet role-sensitive scaffolds critically shape play. For example, NPC quest givers may benefit from rigid symbolic boundaries, while improvisational “hint givers” may require openness to sustain engaging, adaptive dialogue.

We investigated this tension through \textit{The Interview}, a voice-based detective game powered by three GPT-4o NPCs. In a within-subjects usability study ($N=10$), players experienced both highly constrained and minimally constrained prompts. Despite technical differences, they reported no systematic improvement in dialogue quality beyond sensitivity to surface-level breakdowns such as latency or contradictions. Guided by this insight, we redesigned the highly constrained prompt into a hybrid \texttt{JSON+RAG} architecture, encoding symbolic \texttt{JSON} schemas to stabilize structure while preserving improvisational freedom. A synthetic evaluation revealed role-specific trade-offs: \texttt{JSON+RAG} scaffolds improved interviewer (i.e., quest giver NPC) consistency but reduced the two suspects’ (i.e., improvisational NPCs) believability.

This paper makes three contributions to HCI and game research:
\begin{enumerate}
    \item \textbf{Demonstrate empirically} that increasing prompt complexity does not straightforwardly improve player experience; refinements often remain invisible and surface only as role-dependent trade-offs.
    \item \textbf{Introduce Symbolically Scaffolded Play}, a framework that formalizes prompts as fuzzy boundaries: stabilizing coherence where breakdowns disrupt believability, while leaving room for improvisational richness.
    \item \textbf{Establish a methodological blueprint} for generative system evaluation, combining usability studies with synthetic LLM-based probes to reveal what matters to players.
\end{enumerate}

\section{Related Works}
\subsection{Game Design and Procedural Content Generation}
Procedural content generation (PCG) has long been explored in game research as a way to reduce authorial burden, expand replayability, and inject surprise into play \cite{maleki2024pcg}. Classic taxonomies distinguish between representational levels such as bits, spaces, and scenarios, clarifying what is generated and how it structures experience \cite{5756645}. Across this tradition, however, PCG has often been framed as a question of automation, particularly on how much of the designer’s labor can be outsourced to machines.

Large language models (LLMs) shift this emphasis from automation to generative co-creativity. Systems such as QuestGPT \cite{vartinen2022generating} and GENEVA \cite{10645625} show that LLMs can rapidly produce quests and branching narratives, yet they may also expose persistent problems: repetition, incoherence, and cultural bias \cite{al2023questville, taveekitworachai2023journey}. New evaluation frameworks, such as the PCG Benchmark \cite{khalifa2025pcgbenchmark}, highlight trade-offs between diversity, quality, and controllability. Recent conversational PCG systems \cite{whitehead2025conversational} and constraint-driven architectures \cite{xu2025constraint} similarly balance expressive freedom with structural fragility.

Together, this body of work demonstrates that LLM-driven PCG has been celebrated for what it can produce, but less often interrogated for how players actually experience these outputs. Our study takes up this challenge, reframing PCG not as an automation problem but as a usability problem: how scaffolding strategies shape engagement, believability, and coherence in play.

\subsection{LLMs in Games and NPC Dialogue}
If PCG research has emphasized content, work on NPC dialogue has explored how LLMs animate characters. Early studies in interactive fiction \cite{cote2018textworld, yao2020keep} demonstrate that LLMs can expand action spaces by generating plausible continuations beyond rigid command grammars. More recent work equips NPCs with affective states \cite{marincioni2024effect} or hybrid reasoning architectures \cite{zhou2023dialogue}, showing that generative NPCs can evoke lifelike affect and support emergent social learning.

Yet novelty is shadowed by fragility. Players describe LLM-driven NPCs as engaging and surprising, but they also quickly notice contradictions and breakdowns \cite{akoury2023framework, gao2023turing}. Prior studies consistently identify NPC dialogue as both the most promising and the most problematic use of LLMs in games \cite{yang2024gpt, gallotta2024large}. Emerging systems attempt to stabilize this tension: \textit{Slice of Life} uses symbolically grounded states to anchor dialogue without suppressing improvisation \cite{treanor2025slice}, while \textit{Closer Worlds} reframes breakdowns as aesthetic opportunities \cite{lee2024closer}.

What remains underexplored is how players themselves adjudicate this balance. Prior work shows what NPCs can say, but not when or why players forgive errors or celebrate improvisation. Our work addresses this gap directly by testing how different scaffolding strategies shape believability and variation in practice.

\subsection{Prompting Strategies and NPC Design}
Prompting has emerged as the most immediate lever for controlling LLMs. Foundational work demonstrated that careful prompting can elicit emergent reasoning \cite{brown2020language, wei2022chain}, while sampling strategies improve consistency \cite{wang2022self}. Within games, prompts have been used to generate quests \cite{van2021fine}, stories \cite{akoury2023framework}, and NPC roles \cite{gao2023turing}. The pattern is consistent: more rigid prompts reduce contradictions but at the expense of improvisational variety.

Scaffolding techniques extend these strategies by adding structure beyond text-based prompts. \texttt{JSON} schemas specify output formats and knowledge domain boundaries, reducing ambiguity in how models interpret tasks \cite{okuda-askit-2024, marcinkowski-wot-2025}. Retrieval-augmented generation (RAG) complements this by grounding dialogue in external knowledge bases, improving factual consistency and continuity across interactions \cite{lewis2020retrieval, izacard2020leveraging, hu2024game}. Together, \texttt{JSON} and \texttt{RAG} operate as hybrid scaffolds: schemas enforce structural boundaries, while retrieval injects situational memory and lore. Prior work shows such combinations stabilize coherence where breakdowns would be disruptive, though they can also suppress spontaneity in roles that thrive on improvisation \cite{10.1145/3723498.3723702, nasir2023npc, nasir2023practicalpcglargelanguage}.

Memory-based systems likewise aim to preserve continuity, yet often risk repetition \cite{al2023questville}. Even outside games, users face a “prompt literacy gap” that limits their ability to exploit structured strategies \cite{zamfirescu2023johnny}. Recent work therefore emphasizes selective deployment: symbolic scaffolds should be applied where coherence matters most, while leaving space for variation where surprise sustains engagement \cite{figueiredo2025designing, figueiredo2025fuzzy}. Our study builds directly on this trade-off, empirically testing how hybrid prompting strategies—combining symbolic scaffolds with improvisational space—shape player perception across roles.

\subsection{Fuzzy–Symbolic Scaffolding}
Fuzzy logic refers to a computational approach where reasoning is expressed in degrees of truth rather than binary categories, enabling models to capture graded, context-dependent judgments (e.g., ``partially relevant'' rather than strictly true/false) \cite{zadeh-1994-soft}. In the context of machine learning, fuzzy logic is often used to modulate outputs with interpretable rules that accommodate uncertainty and ambiguity, complementing symbolic or statistical methods \cite{mendel2021critical}. Fuzzy logic models graded, context-sensitive reasoning \cite{zadeh-1994-soft, MendelJ.M.1995Flsf}, yet LLMs lack native mechanisms for such modulation. Recent frameworks therefore combine fuzzy rules with symbolic states, offering interpretable anchors while tolerating ambiguity \cite{griol-2021-adaptive, 10469524}.

Prior work propose two such frameworks. One integrated boundary prompts with fuzzy control schemas inspired by Vygotsky’s Zone of Proximal Development, enabling adaptive support without fine-tuning \cite{figueiredo2025designing}. The other paired symbolic scaffolding with short-term memory, showing that removing either degraded abstraction, probing, and conversational continuity \cite{figueiredo2025fuzzy}. Both studies converged: architectural scaffolds reliably shape emergent model strategies, improving alignment and adaptivity.

Recent research reinforces these insights. Treanor et al. \cite{treanor2025slice} demonstrate that symbolically grounded scaffolds stabilize improvisation in social simulations. Xu \& Verbrugge \cite{xu2025constraint} show how constraint-driven representations balance creativity with functional playability in 3D generation. Poglitsch et al. \cite{10.1145/3723498.3723702} modularize scaffolds into memory and planning, showing both their power and their fragility.

What is missing is an empirical, player-centered evaluation of scaffolding strategies. Our work addresses this by testing how scaffolding affects what players notice, value, and enjoy, extending fuzzy–symbolic scaffolding into games.

\subsection{HCI \& Usability in AI Systems}
HCI research reminds us that technical improvements matter little if systems remain unusable or misaligned with user expectations. Amershi et al. \cite{amershi2019guidelines} codify design guidelines for human–AI interaction, highlighting clarity and graceful error recovery. Studies of co-creativity show that AI can inspire but also undermine agency if too prescriptive \cite{lee2024closer}. Recent work on educational games similarly shows how natural-language scaffolds empower non-experts but falter when too rigid \cite{fu2025cracking}.

Games extend these lessons. Conversational PCG \cite{whitehead2025conversational} illustrates how natural-language interaction supports creators while revealing trade-offs between symbolic scaffolds and improvisation. Similarly, LLM-integrated systems such as \textit{LIGS} \cite{jeong2025ligs} highlight how too little scaffolding produces fragmented experiences. Across contexts, usability studies emphasize that players judge systems by whether interactions feel coherent, adaptive, and rewarding \cite{marincioni2024effect, ai6050093}.

Our study advances this line of research by combining usability testing with synthetic evaluation to reveal role-dependent trade-offs in LLM-driven play. This positions usability not as an afterthought to technical performance, but as the central criterion for prompt design.

\section{Game Prototype: The Interview}
\begin{figure}
  \includegraphics[width=\textwidth]{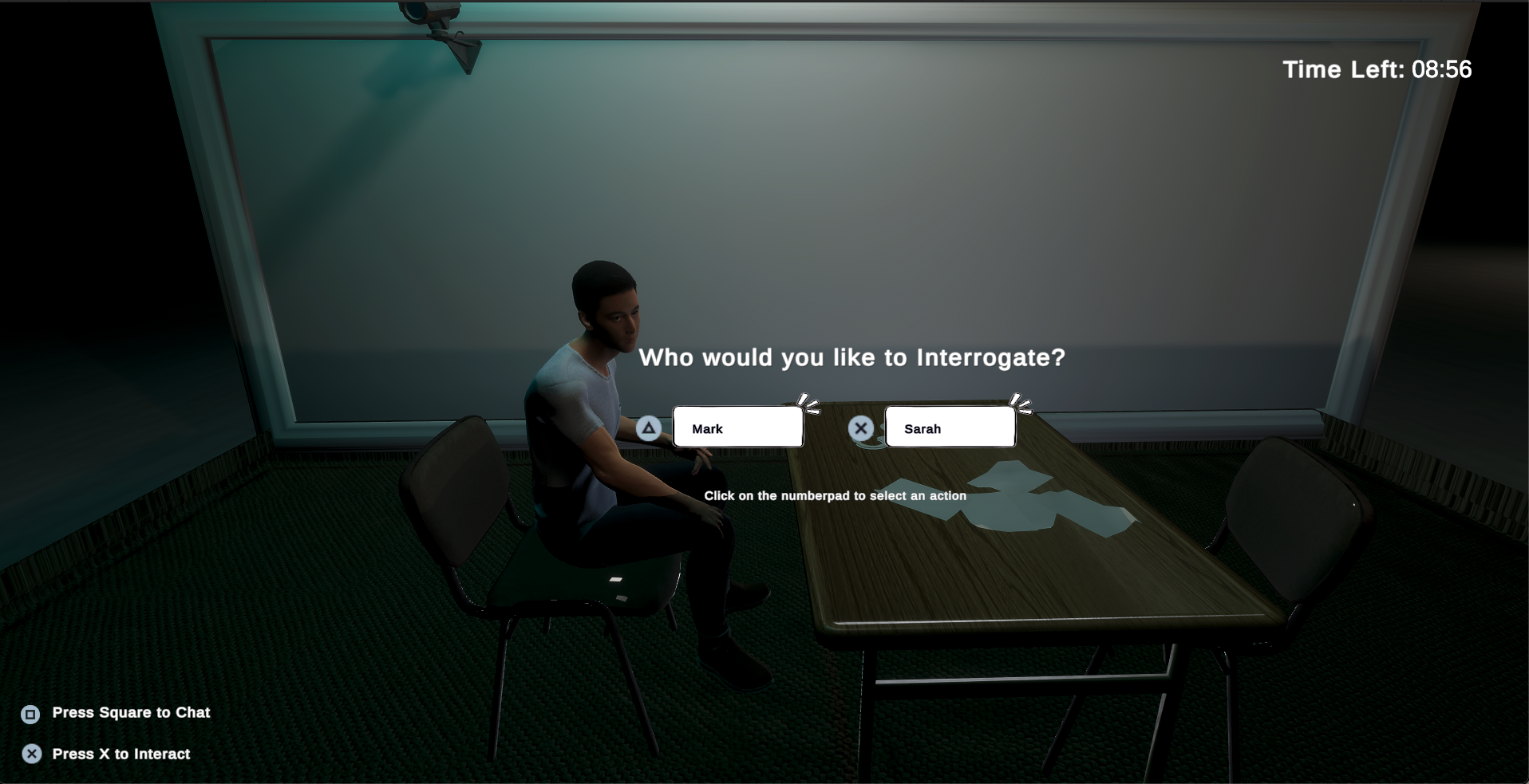}
  \caption{Our prototype, The Interview, demonstrates how role-sensitive scaffolding of LLM prompts can balance coherence and improvisation in NPC dialogue, offering a research probe into Symbolically Scaffolded Play.}
  \Description{Screenshot of The Interview game, showing the player’s detective avatar in a crime-scene office while conversing with an NPC suspect. On-screen subtitles display dialogue, and minimal HUD elements include a timer and speech transcription.}
  \label{fig:game}
\end{figure}

We developed \textit{The Interview} (Fig. \ref{fig:game}), a detective role-playing game designed as a research probe into how LLM-powered NPCs shape player experience in dialogue-driven play. Rather than aiming for a polished entertainment product, our goal was to build a deliberately bounded but functional game world that allowed us to vary scaffolding strategies while preserving ecological validity. Players assume the role of a detective candidate undergoing a job interview. Within this framing, they interrogate two suspect NPCs (Sarah and Mark) while being monitored by an Interviewer (quest-giver NPC), their prospective partner. All three NPCs are powered by GPT-4o, enabling unscripted, real-time dialogue.

\subsection{Gameplay}
\subsubsection{Game Lore and Narrative World}
Set in a near-futuristic Western Canadian city, the \textit{Interview} places players in the role of a homicide detective candidate undergoing a job interview. The game unfolds within a simulation that reconstructs a crime scene, with the player’s performance judged by their ability to communicate, interrogate, and reason. This framing serves both narrative and research purposes, situating the gaming experience as an evaluation of investigative competence.

The narrative world was developed iteratively, beginning with the crime scene lore, organizational backstory, character psychology, a detailed crime timeline, and a final police report. We selected a murder mystery because it naturally supports high-stakes dialogue, complex alibis, and genuine intrigue. The lore centers on employees of the fictional Synergy Solutions, and their relationships, personalities, and actions leading to the crime. This scaffold provides LLMs with a stable foundation of facts, motivations, and interdependencies, enabling improvisations that remain coherent and believable.

\subsubsection{NPCs}
The game features three primary NPCs:

\begin{itemize}
    \item \textbf{The Interviewer (quest-giver):} A senior detective who acts simultaneously as partner, evaluator, narrative anchor, and quest giver. The Interviewer maintains control, probes the player’s reasoning, and delivers key case information. Designed to be rigid and fact-driven, this NPC stabilizes the flow of play.
    \item \textbf{Suspect 1 (Sarah):} The true perpetrator who manipulates her colleague Mark into confessing. Sarah is evasive and emotionally layered, designed to test whether improvisational dialogue can sustain deception, variation, and narrative plausibility.
    \item \textbf{Suspect 2 (Mark):} Sarah’s lover who falsely confesses. Cooperative yet contradictory, Mark’s role requires coherence under pressure, balancing adherence to a false narrative with the emotions of an innocent person accused.
\end{itemize}

\subsubsection{Player Role}
Players act as detective candidates whose task is to interrogate suspects, collaborate with the Interviewer, and assemble a final report. The game's core mechanic is a conversational puzzle. Narrative success does not rely on uncovering a single correct answer but on demonstrating investigative reasoning.

\subsubsection{Player Outcomes}
By the end of play, players are expected to have uncovered the underlying secrets of the company, tested the credibility of conflicting stories, and impressed the Interviewer. Because dialogue is dynamically generated, the mystery unfolds differently across sessions, producing high replayability. The design deliberately emphasizes a trade-off: player success depends less on the objective accuracy of LLM outputs and more on their perceived believability and capacity to sustain engagement.

\subsection{System Architecture and Design Considerations}
A core feature of the prototype is its speech-based interaction loop: players speak naturally, their input is transcribed via speech-to-text (STT), and NPCs respond with generated dialogue delivered through text-to-speech (TTS). This conversational rhythm positioned dialogue as both the mechanic for solving the mystery and the medium for impressing the Interviewer, blending investigation with performative conversational play.

All source code, prompts, and study materials are available as supplemental material.

\subsubsection{Core Architecture in Unity}
The game was built in Unity as a single scene containing all explorable crime-scene environments. To manage system complexity, we implemented a modular architecture with manager scripts for audio, dialogue, LLM interaction, mechanics, and UI. Prompts were stored as \texttt{.txt} or \texttt{.json} files, and at the start of each session the system loaded the appropriate condition (high-constraint prompt---HCP or low-constraint prompt---LCP) and assigned prompts to their respective controllers. The heads-up display (HUD) was intentionally minimal, limited to a timer and NPC subtitles.  

\textbf{Speech Pipeline.} For speech-to-text (STT), we selected Azure Cognitive Services Speech due to cost, ease of integration, and reliability. Azure’s free education tier and better transcription accuracy during pilot testing outperformed alternatives such as OpenAI’s STT. Its integration also provided more granular control over how speech was captured. For text-to-speech (TTS), we used OpenAI voices, which we judged to be more natural and engaging. The main challenge was latency. To address this, we implemented a streaming and queuing pipeline. As soon as the LLM generated a sentence fragment (e.g., ending with a period), it was sent to TTS and added to a playback queue. This allowed NPCs to begin speaking before the full response was complete, significantly reducing perceived lag.  

\textbf{Game Logic.} NPCs were represented as 3D models whose behavior was governed by a central dialogue controller. Turn-taking was mediated by STT, which used extended pauses to signal when a player had finished speaking, after which the NPC would generate a response. The system was designed to be reactive: NPCs responded only when prompted, preventing unnatural over-generation.  

\textbf{Iterative Refinements.} The pilot study revealed that players were uncertain whether NPCs were ``listening.'' In response, we added a UI indicator showing when speech input was active and a live transcription window. These additions improved transparency, reassuring players that their input was captured and correctly processed.  

\subsubsection{Lore Generation}
The lore document was the backbone of character believability. We developed it using an iterative prompt-chaining method with Gemini 2.5 Flash, which consistently produced more coherent scaffolds than OpenAI’s models in pilot testing. Across extensive trials with GPT-4, GPT-4o, and GPT-4o mini, OpenAI models often failed to generate an engaging or coherent narrative when given the same prompts. By contrast, Gemini 2.5 yielded narratives that were more consistent and better aligned with our design goals. For this reason, Gemini 2.5 was used exclusively for lore creation, while OpenAI models were reserved for real-time game prompting. This separation ensured that world-building and gameplay interaction remained distinct processes.  

The lore generation scaffolded in stages:
\begin{enumerate}
    \item High-level scaffolding of game rules and mechanics,
    \item Core murder mystery (crime scene details),
    \item Organizational context of the detective agency,
    \item Character psychology (cognition, motives, dispositions),
    \item Perceptual attributes (relationships, histories, speech styles),
    \item Timeline of events, from weekly to hourly sequences,
    \item Case report written as a third-person police reconstruction.
\end{enumerate}

We separated offline lore-building from online interaction to avoid coupling prompt dynamics to story quality; the lore corpus was held constant across conditions.

This staged approach illustrates how scaffolding in generative games is not reducible to a single prompt but is instead a layered design process: moving from structural anchors (rules, organizations) to fine-grained expressive cues (speech style, relationships).

\subsubsection{Prompt Design}
To examine how prompt specificity influences player experience, we created two distinct prompting strategies used in the usability test: \textbf{High-Constraint Prompt (HCP)} and \textbf{Low-Constraint Prompt (LCP)}.

The HCP version embedded detailed symbolic scaffolds within the system instruction. It included explicit character traits, narrative rules, and behavioral constraints, often specifying dialogue style, role responsibilities, and interaction boundaries. The goal was to maximize coherence and minimize contradictions, ensuring that NPCs consistently adhered to their narrative functions (e.g., suspects maintaining secrecy, Interviewer guiding play). This prompt was paired with the full \texttt{JSON} lore document at runtime, producing a tightly controlled conversational scaffold (Listing \ref{lst:npc_json}). This file included:

\begin{itemize}
    \item character traits, personalities, and speech patterns,
    \item histories and relationships between characters,
    \item crime scene details, timelines, and organizational background,
    \item interaction rules and memory scaffolds.
\end{itemize}

\begin{lstlisting}[language=json, caption={Excerpt of NPC schema for a suspect NPC.}, label={lst:npc_json}]
{
  "suspect_1": {
    "name": "Mark Olsen",
    "emotional_state": {"current": "anxious",
      "transitions": [{"to": "defensive", "trigger_keywords": ["you're lying", "confess", "Sarah did it"]}, {"to": "remorseful", "trigger_keywords": ["you didn't mean to", "you tried to help", "you were manipulated"]}]},
    "cooperativeness": {"current": "low", "transitions": [{"to": "medium", "trigger_keywords": ["I want to help", "you're not alone", "help us understand"]}, {"to": "high", "trigger_keywords": ["we know Sarah's role", "you can clear your name", "we believe you"]}]}
  }
  \end{lstlisting}

Due to space constraints, further details on how the \texttt{JSON} schemas were constructed are provided in the supplementary materials. 

At runtime, the system constructed a context window for GPT-4o that combined relevant lore excerpts, character-specific prompts, dialogue history, and the player’s most recent input (via STT). The model’s responses were then delivered via TTS. While this architecture enabled richly grounded role-play, it exposed a central tension: more contextual scaffolding improved coherence but increased latency ($\sim$3s) and vulnerability to STT errors. These breakdowns shaped later design decisions and motivated our shift toward hybrid symbolic scaffolds.

In contrast, the LCP provided only high-level role descriptions and minimal behavioral guidance, leaving greater latitude for improvisation and variation in NPC dialogue. In this condition, the behavior-specific \texttt{JSON} schema (Listing~\ref{lst:npc_json}) was not loaded, reducing structural constraints on generation.

Together, the two prompts illustrate a central design trade-off: HCP privileges reliability and coherence, whereas LCP emphasizes openness and improvisation.

\subsection{Usability Study: High-Constraint vs. Low-Constraint Prompt}
We conducted a pilot study to refine protocols followed by a main usability study. Both studies were approved by the University Ethics Review Board.

\subsubsection{Pilot Study}
We conducted a pilot study with five participants (female=3; male=2) recruited through the second author's professional network using convenience sampling. Three participants played individually, while two played together as a pair, allowing us to probe whether solo versus paired play influenced perceptions of usability and engagement.

Participants were given verbal instructions to explore the game environment, interact with the NPCs, complete the game’s central objective, and then provide open-ended feedback on their experience.

Findings from the pilot highlighted a key design limitation: opportunities for meaningful interaction with the LLM-powered NPCs were disproportionately concentrated near the end of the session. As a result, much of the early gameplay felt underutilized, limiting the potential for dynamic, conversational engagement throughout. This insight directly informed subsequent revisions, including restructuring the narrative to foreground interrogation earlier and ensuring that conversational play remained the core mechanic across the full session. 

Based on this feedback, we implemented two critical revisions before the main study:
\begin{enumerate}
    \item Increased LLM Interaction. The narrative was restructured around a detective case scenario in which players actively interrogate suspects and share reasoning with their detective partner. This repositioned conversation as the central mechanic.
    \item Transition to Speech Input. The original keyboard interface was replaced with voice input via speech-to-text, enabling more naturalistic and immersive interactions.
\end{enumerate}

\subsubsection{Participants}
We recruited 10 participants (Table \ref{tab:participants}) via classroom announcements and email invitations sent via the university mailing list. All participants reported prior experience with games. Each received a \$25 gift card honorarium. While modest in size, this sample aligns with common practice for formative usability probes in HCI \cite{10.1145/169059.169166}, where the goal is not statistical generalization but to surface design-relevant insights.

\begin{table}[h]
\centering
\caption{Participant demographics and prototype order.}
\Description{This table outlines role-specific scaffolding in The Interview. Symbolic schemas define structural boundaries, while fuzzy logic parameters (0.0–1.0) adapt dynamically to player input and persist in shared memory. This combination creates role-sensitive, computable boundaries that preserve believability while sustaining improvisational freedom.}
\label{tab:participants}
\begin{tabular}{llllll}
\toprule
\textbf{Code} & \textbf{Prototype Order} & \textbf{Age Range} & \textbf{Gender} & \textbf{Self-reported play frequency} \\
\midrule
P01 & HCP $\rightarrow$ LCP & 18--24 & Female & Rarely \\
P02 & LCP $\rightarrow$ HCP & 18--24 & Female & Once a month \\
P03 & HCP $\rightarrow$ LCP & 18--24 & Female & Once a week \\
P04 & LCP $\rightarrow$ HCP & 18--24 & Female & Several times a week \\
P05 & HCP $\rightarrow$ LCP & 35--44 & Female & Rarely \\
P06 & LCP $\rightarrow$ HCP & 25--34 & Male & Rarely \\
P07 & HCP $\rightarrow$ LCP & 18--24 & Male & Once a month \\
P08 & LCP $\rightarrow$ HCP & 25--34 & Male & Several times a week \\
P09 & HCP $\rightarrow$ LCP & 18--24 & Male & Once a month \\
P10 & LCP $\rightarrow$ HCP & 18--24 & Male & Once a month \\
\bottomrule
\end{tabular}
\end{table}

\subsection{Experimental Conditions}
To probe the role of prompt design, we implemented two versions of the system that differed in the specificity of scaffolding:
\textbf{Prototype A: High-Constraint Prompt.} This version provided GPT-4o with a detailed prompt encoding character traits, behavioral rules, and narrative constraints, aiming to tightly control NPC behavior and maximize coherence.
\textbf{Prototype B: Low-Constraint Prompt.} This version offered only high-level descriptions of character roles and personalities, granting GPT-4o more freedom to interpret behavior and generate dialogue.

\subsection{Procedure}
We conducted the usability study in a dedicated usability lab at a Western Canadian university. The setup consisted of a MacBook Air M4 connected to a 32" external monitor, a PlayStation~5 controller, keyboard, and mouse. We selected the controller to approximate the embodied play style of console-based narrative games, where dialogue-driven interaction is typically mediated through handheld devices. Although our prototype ran on a laptop, this choice allowed us to reproduce the tactile rhythm and physical ergonomics of console play without the need to build a dedicated system. All participants used the controller (with mouse and keyboard available only as a fallback), ensuring consistent interaction across sessions and maximizing ecological validity within the constraints of a research prototype. This choice deliberately balanced realism with lab feasibility, ensuring that results would generalize to typical console-style play while remaining tractable in a controlled research setting.

Moreover, controller-based interaction reduces the cognitive load often associated with keyboard-and-mouse setups, particularly for novice players. Prior work shows that familiarity with input devices affects fluency, comfort, and overall usability, with many beginners finding gamepads more approachable than multi-key interfaces \cite{chen2024effect}. By standardizing on a widely used controller, we sought to minimize barriers for participants with varying gaming experience while preserving ecological validity of play.

A microphone was provided to capture player speech for communication with the LLM, while speech output from the NPCs was played through the MacBook Air speaker so that both participants and the researcher could hear responses. To capture both performance and reflection, we used OBS Studio to record participants’ on-screen activity and audio during their think-aloud protocols. The research team also took observational notes to capture moments of breakdown, hesitation, or emergent strategies. We explained the procedure, obtained informed consent, provided instructions on gameplay and controller use, and then stepped outside the booth while remaining available to assist if necessary. Participants could request assistance at any time.

Each participant completed two back-to-back play sessions (approximately 30 minutes each), separated by a 5-minute break.  Prototype order was fully counterbalanced, with five participants beginning in the HCP condition and five in the LCP condition. The conditions were administered single-blind, such that players were unaware of the manipulation (Table~\ref{tab:participants}). This single-blind design minimized expectancy effects and demand characteristics, strengthening the internal validity of comparisons between conditions. This within-subjects design allowed us to compare ratings directly while controlling for individual differences in gaming experience, narrative preference, or familiarity with LLM-based systems.

During gameplay, participants were instructed to:

\begin{enumerate}
    \item Listen to the interviewer's opening dialogue
    \item Interrogate both suspects (Sarah and Mark)
    \item Share reasoning and emerging theories with the interviewer
    \item Reach a conclusion about the crime
\end{enumerate}

After each session, participants completed a 28-item questionnaire (5-point Likert, 1=Strongly Disagree–5=Strongly Agree) covering four constructs: NPC believability (8 items), engagement and immersion (8), behavioral consistency (6), and technical friction (6; e.g., latency, STT/TTS accuracy). Scales were adapted from established game engagement and playability instruments \cite{vanden2020development, brockmyer2009development} but tailored to our research focus on NPC dialogue and user experience. Specifically, we adapted items to foreground the dynamics of LLM-driven dialogue (e.g., believability of spontaneous responses, consistency of suspect alibis, and responsiveness of the Interviewer), aligning general playability constructs with the unique affordances and challenges of generative NPCs. 

In addition, demographic questions (age, gender, and frequency of video game play) were included to contextualize participant backgrounds. Demographics were collected once, immediately after each participant’s HCP session. Because condition order was counterbalanced, this occurred either after the first or the second play session.

\subsection{Results}
\subsubsection{Quantitative Findings}
We compared usability ratings between the High-Constraint Prompt (HCP) and Low-Constraint Prompt (LCP) prototypes in a within-subjects design. Across 28 items, no statistically reliable differences emerged after Holm--Bonferroni correction (all \(p>.30\)). Converging results from two-tailed Wilcoxon signed-rank and paired \(t\)-tests support the interpretation that, under first-play conditions, surface-level breakdowns (latency, clarity, navigation) overshadow hidden prompt scaffolding.

Small numerical trends suggested that HCP encouraged slightly higher reports of engagement and flow (e.g., ``I lost track of time while playing''), whereas LCP was rated marginally higher on adaptability (e.g., ``The game adapted well to my inputs''). Effect sizes were negligible to small (Cliff's $\delta$), and none reached significance.

\begin{table}[h]
\centering
\caption{Largest numerical differences between High-Constraint Prompt (HCP) and Low-Constraint Prompt (LCP). Diff (LCP--HCP): positive values mean LCP scored higher; negative values mean HCP scored higher. None reached statistical significance after correction.}
\Description{This table highlights the largest item-level differences between High-Constraint (HCP) and Low-Constraint (LCP) prompts. While small numerical trends emerged (e.g., HCP rated higher on flow, LCP on adaptability), all differences were negligible to small in effect size and non-significant after correction, reinforcing the overall null finding.}
\label{tab:hcp_vs_lcp_diffs}
\begin{tabular}{p{6.5cm}rrrr}
\toprule
\textbf{Item (abbreviated wording)} & $M_{HCP}$ & $M_{LCP}$ & Diff & Cliff's $\delta$ \\
\midrule
``I noticed repetitive or unnatural content.'' & 3.43 & 2.57 & $-0.86$ & $-0.39$ (small) \\
``I lost track of time while playing the game.'' & 4.10 & 3.67 & $-0.43$ & $-0.20$ (small) \\
``If I made a mistake, I could recover easily.'' & 3.50 & 3.11 & $-0.39$ & $-0.22$ (small) \\
``Error messages and instructions were clear.'' & 3.70 & 3.33 & $-0.37$ & $-0.20$ (small) \\
``The game adapted well to my inputs.'' & 4.22 & 4.56 & $+0.33$ & $+0.21$ (small) \\
``Things seemed to happen automatically.'' & 3.56 & 3.88 & $+0.32$ & $+0.13$ (negl.) \\
\bottomrule
\end{tabular}
\end{table}

When aggregated across all items, HCP again scored only slightly higher ($M = 4.28$) than LCP ($M = 4.07$), with no reliable difference ($p = .36$, $d_z = 0.15$). Taken together, these patterns suggest that the nuances of prompt scaffolding alone were insufficient to produce measurable differences in first-play experiences, pointing instead to the primacy of breakdown recovery and surface-level design cues in shaping usability.

\subsubsection{Qualitative Findings}
Think-aloud protocols, open-ended responses, and field notes revealed four recurrent themes.

\paragraph{Orientation and Usability Challenges.}
First-time players often struggled with navigation and turn-taking, requesting clearer instructions or a tutorial. Several expected the Interviewer NPC to act as a guide.
\begin{quote}\small
\textit{“Not knowing when to click the square button to talk … I constantly overlooked the instructions.”} (P04, LCP)
\end{quote}

\paragraph{Engagement with Dialogue and Roles.}
Dialogue with suspects and the Interviewer was described as the most compelling feature. Improvisation from suspects was valued as lively but risky, while Interviewer consistency was seen as reassuring.
\begin{quote}\small
\textit{“The interactions with Sarah and Mark and also the interviewer were very helpful.”} (P01, LCP)\\
\textit{“I liked the responses of the suspects and how we can ask them questions directly.”} (P03, HCP)
\end{quote}

\paragraph{Requests for Expanded Content and Feedback.}
Players wanted more time, richer evidence interaction, and additional feedback or recap mechanisms.
\begin{quote}\small
\textit{“I enjoyed it so much I wanted more time playing and interacting with the evidence.”} (P01, LCP)\\
\textit{“There should be more mini clip or recap of the scene just before the murder.”} (P09, LCP)
\end{quote}

\paragraph{Cross-Condition Contrasts.}
HCP was described as \textit{“steady”} and \textit{“accurate”} but occasionally “on-rails.” LCP felt \textit{“livelier”} and “more engaging” yet sometimes confusing. These impressions mirror the small, non-significant quantitative differences.

Overall, the findings reinforce a key claim: players rarely perceive the complexity of underlying scaffolds directly. What matters is whether interactions surface in ways that are legible, useful, and motivating. This motivates the principle of \textit{Symbolically Scaffolded Play}: scaffolds should stabilize interactions where breakdowns disrupt play, while leaving space for improvisation where surprise sustains engagement. More broadly, our results highlight a methodological implication for evaluating generative LLM-based systems: usability depends less on hidden prompt sophistication and more on how design surfaces breakdown recovery, onboarding guidance, and role-sensitive improvisation.

\section{Experimental Study: HCP vs. JSON+RAG Prompt Redesign}
\subsection{Experimental Setup}
Building on the usability study, we conducted a synthetic evaluation to probe how redesigned symbolic scaffolds shaped NPC behavior. Specifically, we compared two prompting architectures: (1) the High-Constraint Prompt (HCP) baseline from the usability study, and (2) a redesigned \texttt{JSON+RAG} framework that coupled retrieval-augmented generation with structured state schemas. Because the usability study showed diminishing returns from hidden refinements, we treated synthetic evaluation as a way to 'stress test' scaffolding strategies at scale, probing role-sensitivity in ways impractical for lab studies. 

\begin{figure}
    \centering
    \includegraphics[width=0.9\linewidth]{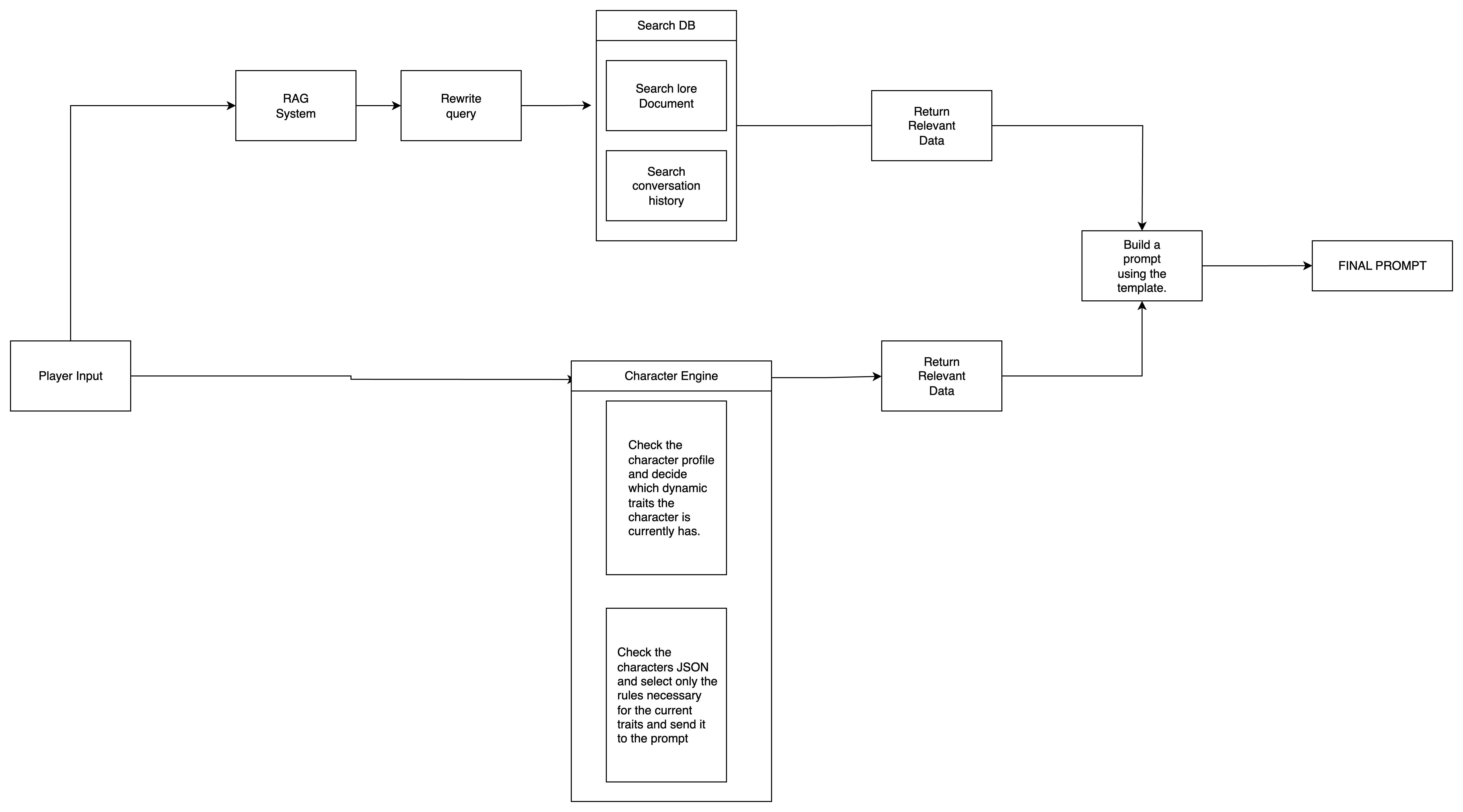}
    \caption{Workflow of the \texttt{JSON+RAG} prompting architecture. Player input is processed through a retrieval pipeline that searches lore and dialogue history, while a character engine selects relevant traits and rules from \texttt{JSON} schemas. Both streams are merged into a structured prompt template, balancing improvisation with symbolic coherence. The workflow illustrates how \texttt{JSON} schemas and retrieval augmentations jointly scaffold NPC behavior, enabling systematic role-sensitive comparisons.}
    \Description{Flowchart illustrating the \texttt{JSON+RAG} workflow. Player input is routed to both a retrieval system, which searches lore and conversation history, and a character engine, which selects rules from character-specific \texttt{JSON} profiles. The outputs are combined into a final structured prompt delivered to the LLM.}
    \label{fig:ragjsonworkflow}
\end{figure}

In the \texttt{JSON+RAG} condition, player input was routed through a RAG pipeline combined with character-specific \texttt{JSON} schemas (Figure~\ref{fig:ragjsonworkflow}). The \texttt{RAG} system first rewrote the query, searched the conversation history and lore database, and returned relevant context. In parallel, a character engine inspected each NPC's dynamic state, selected appropriate traits, and extracted only the relevant rules from its \texttt{JSON} profile. These two information streams were then merged into a prompt template, producing the final structured input to the LLM. This architecture allowed prompts to adapt to narrative context while remaining grounded in symbolic constraints, balancing improvisation with coherence.

\paragraph{LLM judge protocol.}
For the synthetic evaluation, we used an LLM judge (\texttt{gpt-4o} with temperature \(=0.0\)) to rate each output along three anchored 1–4 scales: \emph{Variation} (diversity across two generations of the same prompt), \emph{Relevance} (contextual fit to role and narrative), and \emph{Hallucination} (inverted; higher is fewer contradictions). Each condition was tested across 60 scripted prompt–NPC interactions (20 per role: Interviewer, Sarah, Mark), with two repeated runs per prompt. For every prompt–role pair, each condition (HCP, \texttt{JSON+RAG}) generated two independent responses; the judge scored both and we averaged the pair. To mitigate judge bias, prompts were randomized, labels were neutralized, and temperature was fixed at 0.0; nevertheless, we treat these results as design probes rather than ground truth. While synthetic LLM judging cannot substitute human playtesting, it enables replicable, role-controlled probes of scaffold trade-offs, which we position as a methodological complement to usability testing.

\subsection{Results}

\subsubsection{Aggregated Outcomes}
Aggregated results revealed a consistent pattern: the High-Constraint Prompt (HCP) condition produced higher average scores for \emph{variation} ($M=3.08$ vs.\ $2.83$) and \emph{relevance} ($M=3.28$ vs.\ $2.93$). Hallucination differences were negligible and inconsistent (HCP $M=3.33$ vs.\ \texttt{JSON+RAG} $M=3.17$). Differences in variation and relevance reached statistical significance ($p < .01$, small effects), whereas hallucination effects were negligible. Rather than signaling that one method ``wins,'' these results suggest that scaffolding trade-offs are fundamentally role-contingent: stability in one dimension often comes at the expense of improvisation in another.

\begin{table}[h]
\centering
\caption{Aggregated outcomes across NPCs. Higher scores indicate better performance. Differences suggest that scaffolding influences are not global but role-sensitive.}
\Description{This table reports aggregated evaluation scores comparing High-Constraint Prompting (HCP) with the hybrid JSON+RAG scaffold across all NPC roles. Three metrics were judged by an LLM on 1–4 scales: Variation (diversity across runs), Relevance (contextual fit), and Hallucination (inverted; higher means fewer contradictions). HCP outperformed JSON+RAG on variation and relevance with small effect sizes, while hallucination rates were comparable across conditions. These results suggest that prompt scaffolding effects are not uniformly beneficial but instead highlight role-dependent trade-offs.}
\label{tab:json_results}
\begin{tabular}{lcccc}
\toprule
\textbf{Metric} & \textbf{Condition} & \textbf{Mean} & \textbf{SD} & \textbf{Cliff's $\delta$} \\
\midrule
Variation      & HCP      & 3.08 & 0.28 & $-0.24$ (small) \\
               & \texttt{JSON+RAG} & 2.83 & 0.38 & \\
Relevance      & HCP      & 3.28 & 0.67 & $-0.22$ (small) \\
               & \texttt{JSON+RAG} & 2.93 & 0.25 & \\
Hallucination  & HCP      & 3.33 & 0.75 & $-0.04$ (negl.) \\
               & \texttt{JSON+RAG} & 3.17 & 0.42 & \\
\bottomrule
\end{tabular}
\end{table}

\subsubsection{Role-Specific Effects}
Disaggregating by NPC role revealed the most critical insight: scaffolds succeed or fail depending on narrative function. For the Interviewer---a rule-enforcer and narrative anchor---\texttt{JSON+RAG} produced more stable and predictable outputs, ensuring reliability where contradictions would undermine trust. For Suspects (Mark and Sarah), however, \texttt{JSON+RAG} reduced both variation and relevance (e.g., variation $M=2.75$ vs.\ $3.20$; relevance $M=2.85$ vs.\ $3.50$, $p < .05$, medium effects). Here, improvisation is the core of believability: rigid scaffolds drained spontaneity, making alibis feel less plausible. Suspects, as secondary characters, showed only negligible differences between conditions.

\begin{table}[h]
\centering
\caption{Role-specific trade-offs between High-Constraint Prompting and \texttt{JSON+RAG} Scaffolding.}
\Description{This table shows that scaffolding effects are role-dependent: JSON+RAG improves stability for the Interviewer but reduces improvisation and believability for Suspects, while offering only minor clarity gains for supporting roles.}
\label{tab:role_scaffold_matrix}
\begin{tabular}{p{3.2cm}p{5cm}p{5cm}}
\toprule
\textbf{Role} & \textbf{High-Constraint Prompt} & \textbf{JSON+RAG Scaffold} \\
\midrule
\textbf{Interviewer (quest-giver)} \newline (rule-enforcer, narrative anchor) 
& Varied, sometimes contradictory, but adaptive to context 
& \checkmark\ Stable \& consistent \newline \xmark\ Less flexible \newline Strong for reliability \\
\midrule
\textbf{Suspect 1 (Mark)} \newline (needs alibis, improvisation) 
& \checkmark\ Improvisational, surprising, engaging \newline Risk: incoherence 
& \xmark\ Reduced variety, less believable \newline Weak improvisational freedom \\
\midrule
\textbf{Suspect 2 (Sarah)} \newline (supporting role, filler dialogue) 
& Mixed outcomes; context-dependent 
& Neutral effect; minor clarity gains \\
\bottomrule
\end{tabular}
\end{table}

These findings indicate that prompt architecture cannot be judged in the abstract but must be aligned with role-specific narrative functions. Symbolic scaffolds are most valuable where breakdowns are disruptive (e.g., Interviewer consistency), but overly restrictive where variation and surprise sustain engagement (e.g., Suspect improvisation).

This role-sensitive asymmetry directly advances our framework of \emph{Symbolically Scaffolded Play}: scaffolds should act as fuzzy logic boundaries, stabilizing interactions when failure threatens narrative coherence while leaving room for improvisation where play depends on surprise, particularly when a clear decision-making clear cannot be anticipated. More broadly, these results illustrate a methodological implication for HCI: in evaluating LLM-based systems, usability hinges less on hidden prompt sophistication and more on how designs surface breakdown recovery, role guidance, and improvisational flexibility.

Our combined studies advance a methodological lesson: testing LLM-based systems requires not just more prompt tuning, but role-sensitive scaffold design and hybrid evaluation (usability + synthetic probes). This contributes to HCI’s ongoing effort to develop design-oriented evaluation methods for systems that are too variable to be judged by standard usability metrics alone.

\section{Discussion}
\subsection{Reframing Evaluation Logic in Generative Play}
Our findings overturn the dominant evaluation logic in PCG and NPC research. Instead of asking whether LLMs can generate coherent dialogue, we ask which scaffolds matter to players, and when. Prior work has emphasized controllability, factuality, or diversity as technical metrics of progress \cite{nasir2023npc, nasir2023practicalpcglargelanguage, hu2024game}, yet our results reveal that these refinements often remain invisible. What players actually perceive are surface-level qualities, such as responsiveness, believability, and coherence in the moment. In short, players do not reward hidden refinements; they reward scaffolds that make interaction legible, responsive, and believable.

\subsection{Rethinking Assumptions in PCG and NPC Dialogue}
In the broader PCG tradition, automation has long been framed as a question of “how much” content can be generated \cite{maleki2024pcg, vartinen2022generating}. Similarly, studies of generative NPCs celebrate their novelty but lament their fragility \cite{yang2024gpt, gallotta2024consistent}. Our findings both confirm and complicate this picture. Players noticed breakdowns such as contradictions or latency, consistent with prior work, but they did not perceive incremental improvements from high-constraint prompts. This suggests diminishing experiential returns once a threshold of believability is reached. In other words, refining prompts for diversity or factuality may improve log files but not lived play.

\subsection{Prompting, Scaffolding, and Role-Dependent Trade-Offs}
The experimental study extended these insights by treating synthetic evaluation not as a substitute for user testing but as a design probe. \texttt{JSON+RAG} scaffolds indeed provided greater structural consistency, echoing findings from prior prompting pipelines \cite{hu2024game}. Yet they simultaneously reduced variation and contextual fit in suspect roles, which depend on improvisation and surprise. Conversely, free-text prompts fostered adaptive, varied dialogue but sometimes risked incoherence.

There is no universally optimal scaffold, only scaffolds that fit or misfit the role at hand. Our results show that scaffolding cannot be applied uniformly: the Interviewer (quest-giver) benefit from rigid symbolic constraints that preserve order and narrative continuity, while suspects demand openness that sustains improvisational alibis. This role-specific differentiation sharpens and extends prior accounts of the creativity–reliability trade-off: scaffolding strategies must be selectively aligned to character function in play.

\subsection{Extending Fuzzy–Symbolic Scaffolding into Play}
Fuzzy–symbolic scaffolding \cite{figueiredo2025designing, figueiredo2025fuzzy} has shown that hybrid structures combining symbolic anchors with linguistic flexibility enable adaptive reasoning in dialogue systems. Our findings confirm this principle in a new domain: interactive play. Just as rigid prompts constrain in-context adaptivity in tutoring \cite{figueiredo2025designing}, we demonstrate that overly constrained prompts suppress improvisational engagement in games. Extending this theory, we articulate \textit{Symbolically Scaffolded Play}: symbolic structures should operate as fuzzy boundaries, stabilizing coherence only where breakdowns threaten believability while leaving space for improvisation elsewhere.

\subsection{Implications for HCI and Game Design}
Reframing scaffolding around player perception rather than technical optimization yields three design imperatives:
\begin{itemize}
    \item \textbf{Design for perceptibility.} Refinements matter only if players can feel them. Latency, tone, role adherence, and narrative flow outweigh invisible prompt optimizations.
    \item \textbf{Balance freedom and constraint.} Overconstraint undermines improvisation; underconstraint risks incoherence. Hybrid scaffolds, tuned to role, achieve the best experiential balance.
    \item \textbf{Reposition usability testing.} Combining player-centered usability studies with synthetic evaluations exposes where scaffolds enhance experience and where they do not.
\end{itemize}

This methodological hybrid---iterating between player-centered usability studies and synthetic evaluations---offers a replicable blueprint for evaluating LLM-based systems across domains, especially where system complexity makes purely human studies impractical.

\subsection{Limitations}

\textbf{Study Design and Sample Size.} Our usability study used a within-subjects design with $N=10$, suitable for formative inference but underpowered for detecting small effects across 28 items after multiple-comparison correction. Results should therefore be interpreted as evidence for \textit{no large experiential differences} between HCP and LCP under first-play conditions, not as proof of equivalence.

\textbf{Participant Composition.} The participant pool comprised university students with mixed gaming backgrounds; 40\% ($N = 4$) reported playing video games at least once a month. This skews toward \textit{infrequent to moderate} play. For novice or casual players, consistency can aid orientation more than early personalization; experienced players might value improvisational variety sooner. Consequently, null differences between HCP and LCP may partly reflect first-time novice needs rather than a universal insensitivity to prompt refinements.

\textbf{Onboarding Effects. }All participants experienced a single, $\sim$30-minute session per condition. For first-time play, repetition can help players track tasks, build a mental model, and stabilize expectations; personalized or highly varied responses may become more desirable only after familiarity increases. Our results likely reflect \textit{early-stage learning dynamics}, not long-run preferences.

\textbf{Synthetic Evaluation Constraints. }The experiment used scripted prompts and a \texttt{gpt-4o} judge as an early-stage design probe. While efficient and reproducible, LLM judges can encode model biases (e.g., favoring syntactic regularity), potentially inflating the apparent benefit of structured outputs and under-valuing creative drift. Our judge rubric emphasized variation, relevance, and hallucination; future work should triangulate with human raters and task-grounded performance (e.g., clue retrieval, contradiction spotting).

\textbf{Lore and Pipeline Dependencies. }Our ''high-constraint'' architecture relied on a large \texttt{JSON} lore and dynamic prompts; the \texttt{JSON+RAG} redesign constrained outputs further via symbolic schemas and retrieval. Conclusions about role-dependent trade-offs therefore presume comparable lore quality across conditions. Although held constant at design time, minor retrieval differences or schema fit could differentially shape outputs per role.

\textbf{What These Limits Do and Do Not Imply. }These constraints do \textit{not} undermine our central contribution because it demonstrates that, for first-time play in a dialogue-driven detective game, hidden prompt refinements yielded diminishing experiential returns, and that scaffold value is role-dependent. Rather, they carve out clear next steps: (i) longitudinal or multi-session studies to test when improvisational variety becomes a benefit; (ii) diversified samples stratified by gaming frequency; (iii) human-in-the-loop combined with synthetic evaluation; and (iv) genre-targeted replications where personalized prompting is intrinsic to the aesthetic goal (e.g., intimacy-forward play experiences) \cite{xu2025constraint,lee2024closer}.

\subsection{Symbolically Scaffolded Play}
Our findings reveal a novel perspective: there is no single prompt design that guarantees better play. Prompt refinements may improve internal consistency, but players often cannot perceive these changes and, in some cases, overconstraint actively diminishes improvisation. What matters is how scaffolds are strategically differentiated: sets of fuzzy logic boundaries aligned with each NPC role in the game (e.g., one set of the interviewer and, another for suspects).

This lesson confirms prior work on fuzzy–symbolic scaffolding \cite{figueiredo2025designing, figueiredo2025fuzzy}: modular prompting techniques must be separated from gameplay scripts and encoded in reusable structures (\texttt{.json}) that LLMs can flexibly draw upon.

We position \textit{Symbolically Scaffolded Play} as a generalizable design principle for LLM-based systems, not restricted to games. Wherever improvisation and stability must co-exist, such as tutoring or social simulation, role-sensitive scaffolds can mediate usability. To make this principle more concrete, Table \ref{tab:scaffold_strategies} outlines role-specific scaffolding strategies in \textit{The Interview}. The Interviewer (quest-giver NPC) benefits from rigid symbolic structures that enforce coherence and turn-taking, while suspects (NPCs) require looser scaffolds that preserve improvisational variation.

\begin{table}[h]
\centering
\caption{Role-specific scaffolding strategies in \textit{The Interview}, illustrating \textit{Symbolically Scaffolded Play}. Each fuzzy parameter is expressed as a continuous value between $0.0$ and $1.0$, dynamically updated based on player input and persisted in a shared \texttt{JSON} short-term memory schema. Symbolic schemas define role boundaries; fuzzy logic provides graded modulation of behavior.}
\Description{This table shows role-specific scaffolding strategies in \textit{The Interview}, illustrating \textit{Symbolically Scaffolded Play}. Each fuzzy parameter is expressed as a continuous value between $0.0$ and $1.0$, dynamically updated based on player input and persisted in a shared \texttt{JSON} short-term memory schema. Symbolic schemas define role boundaries; fuzzy logic provides graded modulation of behavior.}
\label{tab:scaffold_strategies}
\begin{tabular}{p{2.5cm}p{4.2cm}p{4.2cm}p{4.2cm}}
\toprule
\textbf{NPC Role} & \textbf{Symbolic Scaffold (JSON Schema)} & \textbf{Fuzzy Logic Parameters} & \textbf{Expected Player Experience} \\
\midrule
\textbf{Interviewer (quest-giver)} & 
Turn-taking controller; recap/agenda templates; contradiction-check and evidence-citation rules; ``no new facts'' and ``no spoilers'' constraints & 
Guidance intensity $g$ ranges from $0.0$ to $1.0$, increasing when players fail to collect evidence or when suspects are evasive; recap frequency $r$ rises as contradictions accumulate. All updates written to shared memory for continuity & 
Dialogue feels stable, focused, and supportive; Interviewer (quest-giver) adapts guidance without revealing solutions \\
\midrule
\textbf{Suspect} & 
Alibi graph (events \& dependencies); deception tactics; emotion-state transitions; ``forbidden facts'' list & 
Evasiveness $e$ ranges from $0.0$ to $1.0$, decreasing with rapport-building inputs; disclosure probability increases when confronted with consistent evidence. Updates stored in shared memory so evasiveness carries over across turns & 
Interactions feel unpredictable yet role-consistent, balancing coherence with improvisation \\
\bottomrule
\end{tabular}
\end{table}

A distinctive feature of our approach is that each NPC’s behavior is defined by \emph{numerical fuzzy logical ranges} that bound the conditions under which the model will reason and respond. Unlike Listing~\ref{lst:npc_json}, which encodes natural-language state descriptors (e.g., ``low'' or ``high'' cooperation), the schema in Table~\ref{tab:scaffold_strategies} specifies continuous values between $0.0$ and $1.0$ for each fuzzy parameter. This design enables more granular modulation of scaffolding: the \textit{Interviewer (quest-giver)}’s guidance intensity, for instance, is anchored between $0.0$ and $1.0$ and dynamically increases if the player fails to gather sufficient evidence from suspects. Conversely, a \textit{Suspect}’s evasiveness is similarly encoded between $0.0$ and $1.0$, rising when the player’s dialogue choices undermine rapport (e.g., by triggering confrontational keywords). 

Crucially, these numerical values are not transient. Each adjustment is stored in persistent gameplay memory, implemented as a shared \texttt{JSON} document—that all NPCs consult to refresh their current state across turns (Listing~\ref{lst:memory_min}).

\begin{lstlisting}[language=json, caption={Abridged shared short-term memory (exemplary). Full schema in Supplementary Materials.}, label={lst:memory_min}]
{"turn_index": 12, "contradiction_count": 2, "recent_evidence": ["ev_017", "ev_021"], "player_rapport": { "with_suspect": 0.42 }, "npc_state": {"interviewer": {"guidance_intensity": 0.58 }, "suspect":     { "evasiveness": 0.63, "disclosure_prob": 0.22 }}, "last_recaps": { "interviewer": 9 }}
\end{lstlisting}

Listing~\ref{lst:memory_min} exemplifies how NPC states and interaction history are persisted during play. Each entry records turn progression, contradictions, and evidence, alongside graded fuzzy parameters (all $0.0$–$1.0$) that modulate behavior: the \textit{Interviewer (quest-giver)} adapts guidance intensity, while the \textit{Suspect} adjusts evasiveness and disclosure probability. By storing these evolving values in a shared memory, NPCs refresh their states across turns, ensuring continuity and role-sensitive adaptation rather than one-off responses. Thus, we establish a mechanism for both interpretability (designers can read and tune the ranges) and adaptability (the model can interpolate and reason within them). 

To further anchor this principle, Listing \ref{lst:scaffolds_min} provides a \texttt{JSON} snippet that encodes fuzzy–symbolic scaffolds for each role. Note how symbolic schemas define role expectations (e.g., \texttt{guidance\_intensity} and \texttt{evasineness}), while fuzzy logic introduces graded variation that sustains believability without collapsing improvisational freedom.

\begin{lstlisting}[language=json, caption={Exemplary role scaffolds with fuzzy ranges. Full schema available in Supplementary Materials.}, label={lst:scaffolds_min}]
{"npc_roles": {
    "interviewer": {"symbolic_schema": {"constraints": ["no_new_facts", "no_spoilers"]}, "fuzzy_params": {"guidance_intensity": { "default": 0.55, "min": 0.0, "max": 1.0 }}, "update_rules": [{"when": { "suspect_evasiveness": ">5" }, "then": { "guidance_intensity": "+0.10" } }]},

    "suspect": {"symbolic_schema": {"constraints": ["forbidden_facts_filtered"]}, "fuzzy_params": {"evasiveness": { "default": 0.55, "min": 0.0, "max": 1.0 }, "disclosure_prob": { "default": 0.25, "min": 0.0, "max": 1.0 }}, "update_rules": [{ "when": { "evidence_count": ">=2" }, "then": { "disclosure_prob": "+0.20", "evasiveness": "-0.10" }}]}
    }
}
\end{lstlisting}

Listing~\ref{lst:scaffolds_min} illustrates how scaffolds in \textit{The Interview} are encoded as \textit{computable ranges} between $0.0$ and $1.0$. For the \textit{Interviewer (quest-giver)}, a guidance intensity parameter rises when suspects are evasive, prompting more support without revealing the solution. For the \textit{Suspect}, evasiveness decreases and disclosure probability increases when the player presents consistent evidence. Each update is written to a shared short-term memory, ensuring that NPC states persist and evolve across turns \ref{lst:memory_min}. 

These schemas exemplify a methodological novelty: \textit{scaffolds as dynamic, numerical boundaries that adapt with play history}, balancing stability with improvisational freedom. Scaffolds become \emph{computable boundaries} that flex with interaction history, stabilizing believability where breakdowns would otherwise occur while preserving improvisational space where variation is essential.

\section{Conclusion}
The future of generative play cannot be reduced to prompt optimization alone. Through a pipeline of usability testing and synthetic evaluation, we showed that players remain largely indifferent to hidden refinements, perceiving value only when coherence and believability visibly break down. Even when redesigned with \texttt{JSON+RAG} scaffolds, prompts produced role-dependent trade-offs: stability for the Interviewer (quest-giver), but suppressed improvisation for suspects. These findings overturn the assumption that tighter constraints necessarily yield better play.

From this, we articulated the framework of \textit{Symbolically Scaffolded Play}. Extending fuzzy–symbolic scaffolding into games, we argue that prompts should function as fuzzy boundaries: symbolic anchors that enforce coherence only where it matters most, while leaving space for improvisation where surprise drives engagement. This reframing positions scaffolding not as a technical patch, but as a core design principle for believable, player-centered generative systems.

The implications reach beyond games. For designers, \textit{Symbolically Scaffolded Play} provides a practical method: modularize scaffolds by role, encode them in declarative files, and let LLMs improvise within fuzzy constraints. For researchers, it offers a methodological contribution: combining usability studies with synthetic evaluations to expose experiential trade-offs that technical benchmarks obscure. This dual contribution—a design principle and a method—establishes a generalizable model for HCI research on generative systems.

More broadly, our work reframes evaluation in HCI. The critical question is no longer Can LLMs produce coherent dialogue? but rather How should we scaffold generative systems so that coherence and creativity become meaningful to people? By centering usability and role-sensitive scaffolding,\textit{Symbolically Scaffolded Play} advances a foundation for designing generative interactions that are not only functional, but experientially compelling.


\bibliographystyle{ACM-Reference-Format}
\bibliography{paper}










\end{document}